\documentclass[11pt]{article}

\usepackage{epsfig}

\begin{document}

%
%


\newcommand{\aven}[0]{AVENTINUS }


\newlength{\oldparskip}
\setlength{\oldparskip}{\parskip} 
\oldparskip=\parskip 

\newcommand{\resetparskip}[0]
{
  \parskip=\oldparskip
}


\newcommand{\bit}[0]
{
  \parskip=-4pt
  \begin{itemize}
  \itemsep=-3pt
}
\newcommand{\eit}[0]
{
  \end{itemize}
  \parskip=-4pt
}


\newcommand{\ben}[0]
{
  \parskip=-4pt
  \begin{enumerate}
  \itemsep=-3pt
}
\newcommand{\een}[0]
{
  \end{enumerate}
  \parskip=-4pt
}


\newcommand{\bde}[0]
{
  \parskip=-4pt
  \begin{description}
  \itemsep=-3pt
}
\newcommand{\ede}[0]
{
  \end{description}
  \parskip=-4pt
}

%
%
\newcommand{\notes}[1]
{
  \mbox{ } \\
  NOTES
  \begin{verse}
  {#1}
  \end{verse}
  ENDNOTES \\
}


\newcommand{\tickle}[0]{Tcl/Tk }
\newcommand{\cpp}[0]{C[++] }

\newcommand{\needed}[2]{=={#1}=={#2}==}

\newcommand{\delete}[1]{}
\newcommand{\omitted}[1]{{\small \sf =OMITTED=}}

\renewcommand{\notes}[1]{}



\begin{titlepage}
\begin{center}

\Huge
Information Extraction --- \\
a User Guide \\

\mbox{ } \\

\large
\vspace*{.5cm}
Hamish Cunningham \\
\mbox{ } \\
January 1997 \\
Research memo CS -- 97 -- 02 \\
\mbox{ } \\
Institute for Language, Speech and Hearing (ILASH), and \\
Department of Computer Science \\
University of Sheffield, UK \\
\mbox{ } \\
h.cunningham@dcs.shef.ac.uk \\
http://www.dcs.shef.ac.uk/research/groups/nlp/extraction/ \\
http://www.dcs.shef.ac.uk/\verb|~|hamish

\end{center}
\end{titlepage}

\pagestyle{myheadings}
\markboth{IE -- A User Guide}
         {IE -- A User Guide}

\renewcommand{\baselinestretch}{0.8}
\small \normalsize
\tableofcontents
\renewcommand{\baselinestretch}{1.4}
\small \normalsize

\bibliographystyle{alpha}
%
%
%

\section{Introduction}

This note gives a user-oriented view of Information Extraction (IE). No
knowledge of language processing is assumed. For a more technical overview
see \cite{Cow96}.

Information Extraction is a process which takes unseen texts as input and
produces fixed-format, unambiguous data as output. This data may be used
directly for display to users, or may be stored in a database or spreadsheet
for later analysis, or may be used for indexing purposes in Information
Retrieval (IR) applications.

It is instructive to compare IE and IR: whereas IR simply finds texts and
presents them to the user, the typical IE application
analyses texts and presents only the specific
information from them that the user is interested in. For example, a user
of an IR system
wanting information on the share price movements of companies with holdings
in Bolivian raw materials would typically type in a list of relevant words
and receive in return a set of documents (e.g. newspaper articles)
which contain likely matches. The user would then read the documents and
extract the requisite information themselves. They might then enter the
information in a spreadsheet and produce a chart for a report or
presentation. In contrast, an
IE system user could, with a properly configured application,
automatically populate their spreadsheet directly with
the names of companies and the price movements.

There are advantages and disadvantages to IE with respect to IR. IE systems
are more difficult and knowledge-intensive to build, and are to varying
degrees tied to particular domains and scenarios (see next section).
They are also
(for most tasks) less accurate than human readers. IE is more
computationally intensive than IR.
However, in applications where there are large text volumes
IE is potentially much more efficient than IR because of the possibility of
reducing the amount of time analysts spend reading texts.
Also, where results need to be presented in several languages, the fixed
format, unambiguous nature of IE results makes this straightforward in
comparison with providing full translation facilities.

\section{\label{types}Types of IE}

There are four types of information extraction (or information extraction
{\em tasks}) currently available
(as defined by the leading forum for this research, the Message
Understanding Conferences \cite{Gri96}.).
\begin{description}
\item[Named Entity recognition (NE)] \mbox{} \\
Finds and classifies names, places etc.
\item[Coreference Resolution (CO)] \mbox{} \\
Identifies identity relations between entities in texts.
\item[Template Element construction (TE)] \mbox{} \\
Adds descriptive information to NE results.
\item[Scenario Template production (ST)] \mbox{} \\
Fits TE results into specified event scenarios.
\end{description}
From a user point-of-view, NE, TE and ST are the most relevant IE tasks (CO,
as noted below, is necessary as an adjunct to the other tasks, but is of
limited direct usefulness to the IE system user). NE, TE and ST provide
progressively higher-level information about texts.

These are described in more detail below, after a discussion of the current
performance levels of IE technology.

\section{\label{performance}Performance levels}

Each of the four types of IE have been the subject of rigorous
performance evaluation in MUC-6 (1995) and other MUCs, 
so it is possible to say quite precisely
how well the current level of technology performs. Below we will quote
percentage figures quantifying performance levels -- they should be
interpreted as a combined measure of precision and recall (see the section
on evaluation in \cite{Arp95}).
Several caveats should be noted: most of the evaluation has been on English
(with some Japanese, Chinese
and Spanish) -- some applications of the technology may be
either easier or more difficult in other languages.

The performance of each IE task, and the ease with which it may be
developed, is to varying degrees dependent on:
\begin{description}
\item[Text type:] the kinds of texts we are working with, for
example Wall Street Journal articles, or email messages, or HTML documents
from the World Wide Web.
\item[Domain:] the broad subject matter of those texts, e.g.
financial news, or requests for technical support, or tourist information.
\item[Scenario:] the particular event types that the IE user is
interested in, for example mergers between companies, or problems
experienced
with a particular software package, or descriptions of how to locate parts
of a city.
\end{description}
For example, a particular IE application might be configured to process 
financial
news articles from a particular news provider and find information about 
mergers between companies and various other scenarios.
The performance of the application would be predictable for
only this conjunction of factors. If it was later
required to extract facts from the love letters of Napoleon Bonaparte as 
published on wall posters in the 1871 Paris Commune, performance levels
would no longer be predictable. Tailoring an IE system to new requirements
is a task that varies in scale dependent on the degree of variation in the
three factors listed above.

\section{\label{ne}Named Entity recognition}
\begin{figure}[!htb]
  \centerline{\psfig{figure=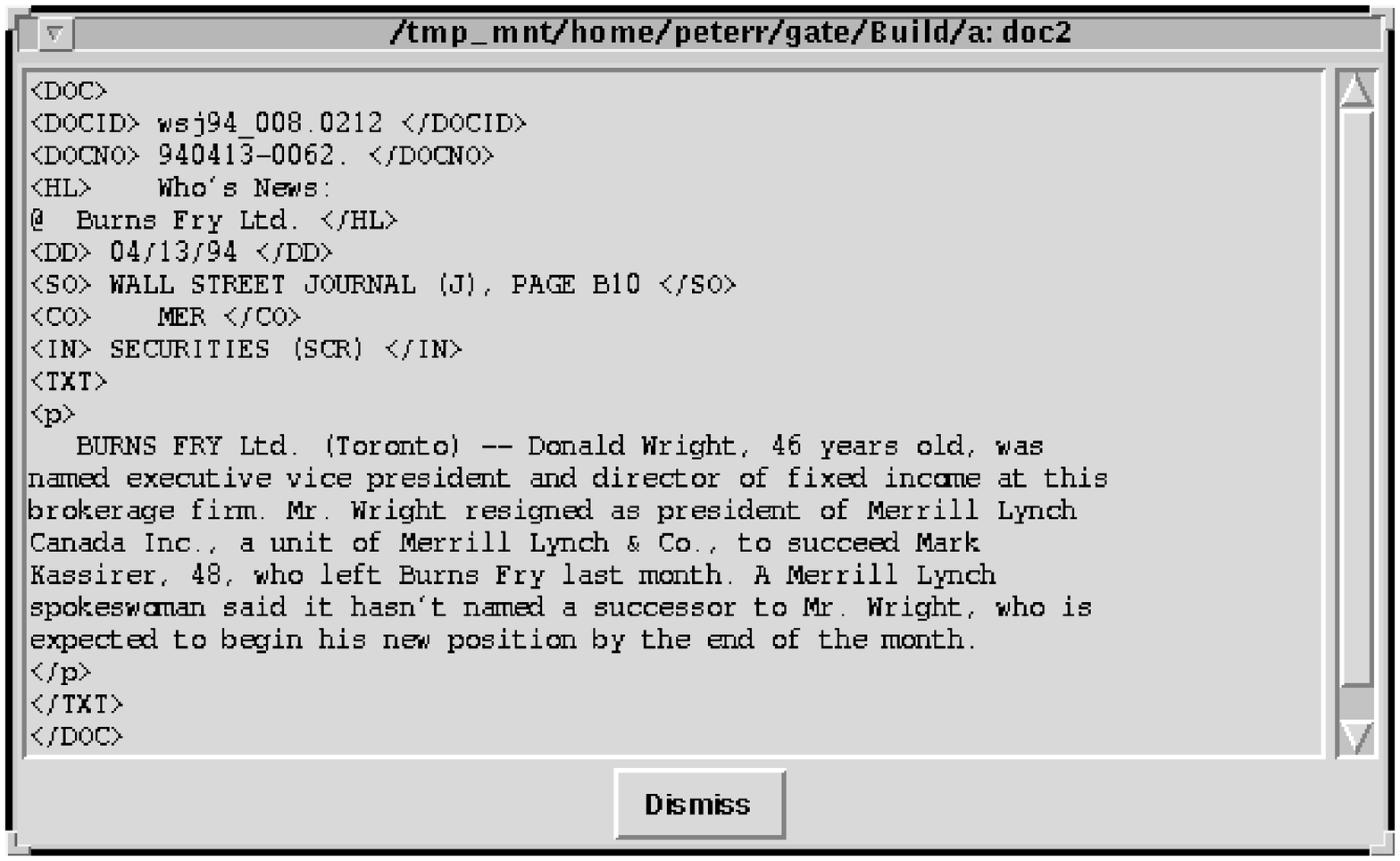,height=8cm}}
  \caption{\label{eg_text}An example text}
\end{figure}

\begin{figure}[!htb]
  \centerline{\psfig{figure=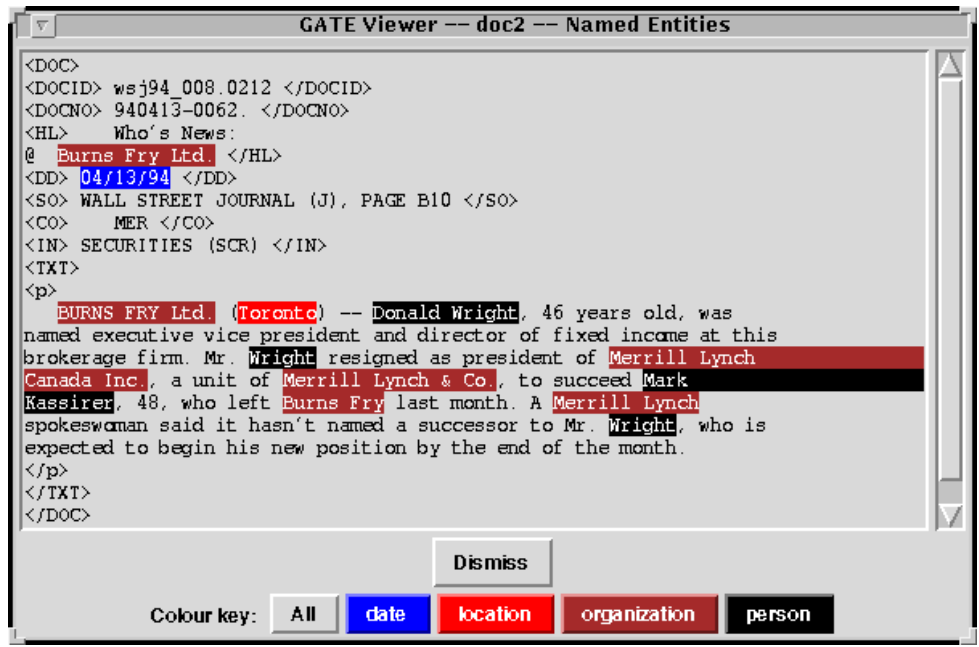,height=8.5cm}}
  \caption{\label{ne_result}Named entity recognition}
\end{figure}

The simplest and most reliable IE technology is {\em Named Entity
recognition} (NE). NE systems identify all the names of people, places,
organisations, dates, and amounts of money. So,
for example, if we run the Wall Street Journal
text in figure \ref{eg_text} through an NE
recogniser, the result is as in figure \ref{ne_result} (this looks better in
colour!). (The viewers shown here and below are part of the GATE language
engineering architecture and development environment -- see \cite{Cun96c}.)
NE recognition can be performed at 96\% accuracy; the current Sheffield
system (\cite{Gai95b})
performs at 92\% accuracy. Given that human annotators do not perform
to the 100\% level (measured in MUC by inter-annotator comparisons), NE
recognition can now be said to function at human performance levels,
and applications of the technology are increasing rapidly as a result.

A recent evaluation of NE for Spanish, Japanese and Chinese (\cite{Mer96})
produced the following scores:

\begin{tabular}{p{1.2in}p{1.2in}p{1.2in}p{1.2in}}
        &       {\bf language}  &       {\bf best system}       &       \\
        &                       &                               &       \\
        &       Spanish         &       93.04 \%                &       \\
        &       Japanese        &       92.12 \%                &       \\
        &       Chinese         &       84.51 \%                &       \\
\end{tabular}

The process is weakly domain dependent, i.e. changing the subject matter of
the texts being processed from financial news to other types of news would
involve some changes to the system, and changing from news to scientific
papers would involve quite large changes.

\section{\label{co}Coreference resolution}

Coreference resolution (CO) involves identifying identity relations between
entities in texts. These entities are both those identified by NE recognition
and anaphoric references to those entities. For example, in
\begin{verse}
Alas, poor Yorick, I knew him well.
\end{verse}
coreference resolution would tie ``Yorick'' with ``him'' (and ``I'' with
Hamlet, if that information was present in the surrounding text).

\begin{figure}[!htb]
  \centerline{\psfig{figure=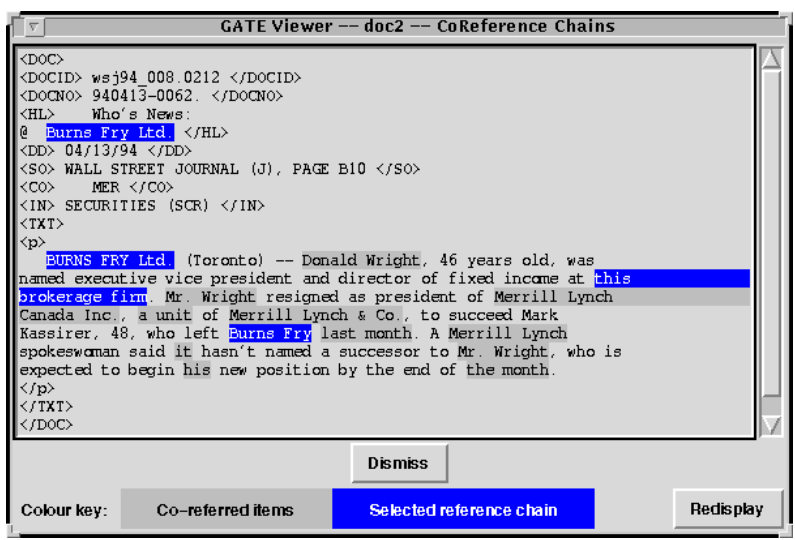,height=7cm}}
  \caption{\label{co_result}Coreference resolution}
\end{figure}

This process is less relevant to users than other IE tasks (i.e. whereas the
other tasks produce output that is of obvious utility for the application
user, this task is more relevant to the needs of the application developer).
For text
browsing purposes we might use CO to highlight all occurrences of the same
object or provide hypertext links between them. CO technology might also be
used to make links between documents, though this is not currently part of
the MUC programme. The main significance of this task, however, is as a
building block for TE and ST (see below). CO enables the association of
descriptive information scattered across texts with the entities to
which it refers. To continue the hackneyed Shakespeare example, coreference
resolution might allow us to situate Yorick in Denmark. Figure \ref{co_result}
shows results for our example text.

CO resolution is an imprecise process when applied to the solution of
anaphoric reference. The Sheffield system scored 51\% recall and 71\%
precision%
\footnote{For statistical reasons the combined precision and recall measure
we use elsewhere is inappropriate here.}
at MUC-6; other systems scored e.g. 59\% recall / 72\% precision,
63\% recall / 63\% precision. These scores are low (although problems with
completing the task definition on schedule complicated matters, and led to
human scores of only around 80\%), but note that this hides
the difference between proper noun coreference
identification (same object, different spelling
or compounding, e.g. ``IBM'', ``IBM Europe'', ``International Business
Machines Ltd.'', \ldots) and anaphora resolution, the former being
a significantly easier problem.

CO systems are domain dependent.

\section{\label{te}Template Element production}

The TE task builds on NE recognition and coreference resolution.
In addition to locating and typing (i.e. classifying, or
assigning to a type -- personal name, date etc.) entities in
documents, TE associates descriptive information with the entities. For
example, from the figure \ref{eg_text} text the system finds out that Burns
Fry Ltd. is located in Toronto, and it adds the information that this is in
Canada.

\begin{figure}[!htb]
  \centerline{\psfig{figure=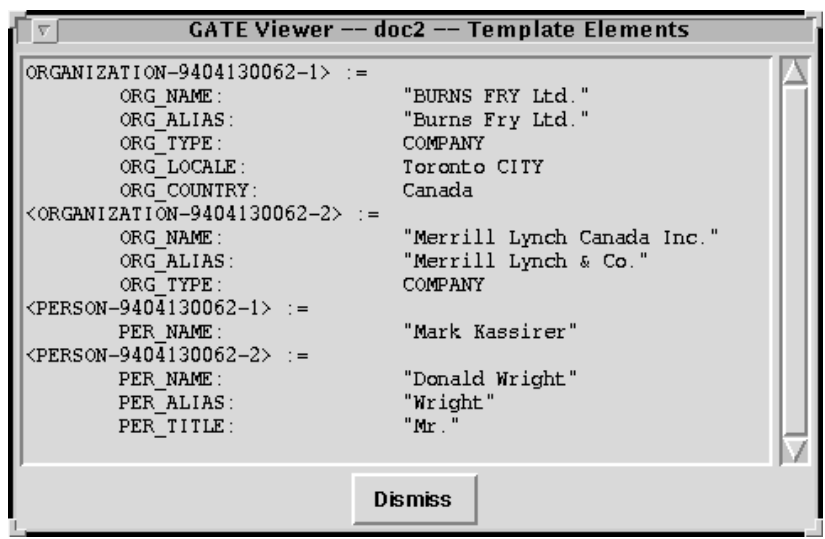,height=7cm}}
  \caption{\label{te_result}Template elements}
\end{figure}

Template elements for the figure \ref{eg_text} text are given in
figure \ref{te_result}.
The format is a somewhat arbitrary one developed at the behest of the
American intelligence community (the original target user group of the MUC
competitions). It is difficult to read; the main point to note is that it
is essentially a database record, and
could just as well be formatted for SQL store operations, or reading into a
spreadsheet, or (with some extra processing) for multilingual presentation.
Section \ref{av_ie} gives a simplified example.

The current Sheffield system scores 71\% for TE production; the best MUC-6
system scored 80\%. Humans achieved 93\%. MUC-6 was the first MUC to
evaluate TE and ST tasks separately -- TE scores should improve in future as
developers gain more experience with the task.

As in NE recognition, the production of TEs is is weakly domain dependent,
i.e. changing the subject matter of
the texts being processed from financial news to other types of news would
involve some changes to the system, and changing from news to scientific
papers would involve quite large changes.

\section{\label{st}Scenario Template extraction}

\begin{figure}[!htb]
  \centerline{\psfig{figure=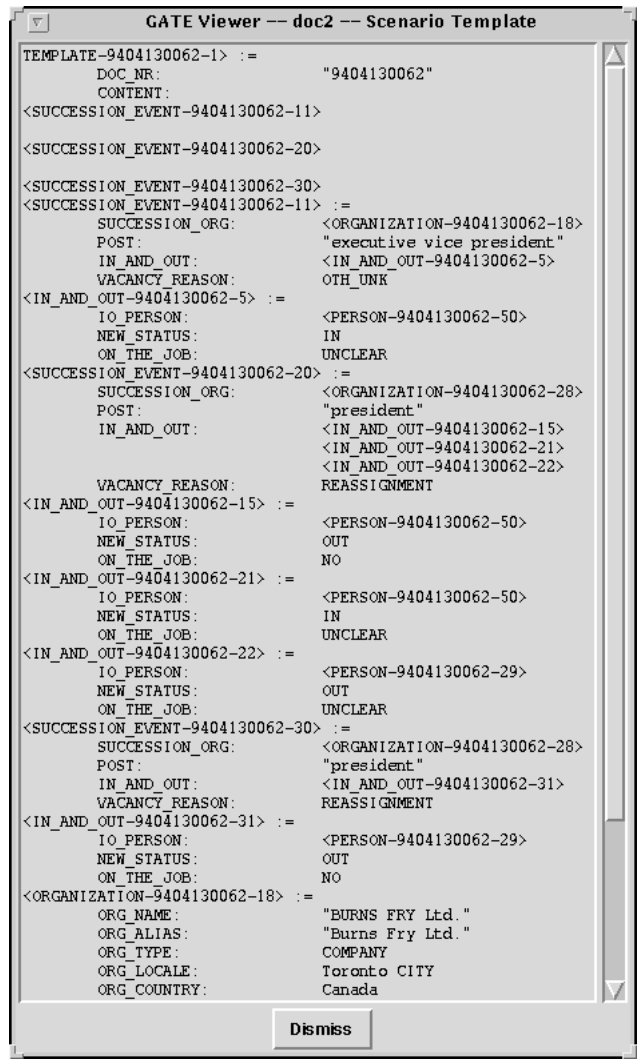,height=14cm}}
  \caption{\label{st_result}Scenario template}
\end{figure}

Scenario templates (STs) are the prototypical outputs of IE systems. They
tie together TE entities into event and relation descriptions. For example,
TE may have identified Isabelle, Dominique and Fran\c{c}oise as people entities
present in the Robert edition of Napoleon's love letters. ST might then
identify facts such as
that Isabelle moved to Paris in August 1802 from Lyon to be nearer to the
little chap, that Dominique then burnt down Isabelle's apartment block and 
that Fran\c{c}oise ran off with one of Gerard Depardieu's ancestors.
A slightly more pertinent example is
given in figure \ref{st_result}.
The same comments regarding format apply as for the TE task.

ST is a difficult IE task.
The current Sheffield system scores 49\% for ST production; the best MUC-6
system scored 56\%. The human score was 81\%, which illustrates the
complexity involved.
These figures should be taken into account when
considering appropriate applications of ST technology. Note, however, that
it is possible to increase precision at the expense of recall: we can
develop ST systems that don't make many mistakes, but that miss quite a lot
of occurrences of relevant scenarios. Alternatively we can push up recall and
miss less, but at the expense of making more mistakes.

The ST task is both domain dependent, and, by definition, tied to the
scenarios of interest to the users. Note however that the results of NE and
TE feed into ST. Note also that in MUC-6 the developers were given
the specifications for the ST task only 1 month before the systems were
scored. This was because it was noted that an IE system that required very
lengthy revision to cope with new scenarios was of less worth than one that
could meet new specifications relatively rapidly. As a result of this,
the scores for ST in MUC-6
were probably slightly lower than they might have been with a longer
development period. Experience from previous MUCs suggests that current
technology has difficulty attaining scores much above 60\% accuracy
for this task, however.

%
%
%
%
%

\section{\label{av_ie}An Extended Example}

So far we have discussed IE from a general perspective. In this section we
look at the capabilities that might be delivered as part of 
an application designed to support analysts tracking international drug
dealing.

When the system is specified, our imaginary analyst states that ``the
operational domains that user interests are centred around are...
drug enforcement, money laundering, organised crime, terrorism,
legislation''. The entities of interest within these domains are cited as
``person, company, bank, financial entity, transportation means, locality,
place, organisation, time, telephone, narcotics, legislation, activity''. A
number of relations (or ``links'') are also specified, for example between
people, between people and companies, etc. These relations are not typed,
i.e. the kind of relation involved is not specified. Some relations
take the form of properties of entities -- e.g. the location of a
company -- whilst others denote events -- e.g. a person visiting a ship.

Working from this starting point an IE system is designed that:
\begin{enumerate}
\item
is tailored to texts dealing with drug enforcement, money laundering,
organised crime, terrorism, and legislation;
\item
recognises entities in those texts and assigns them to one of a number of
categories drawn from the set of entities of interest (person, company,
\ldots);
\item
associates certain types of descriptive information with these entities,
e.g. the location of companies;
\item
identifies a set (relatively small to begin with) of events of interest by
tying entities together into event relations.
\end{enumerate}
For example, consider the following text:
\renewcommand{\baselinestretch}{1.0}
\small \normalsize
\begin{quote}
Reuter -- New York, Wednesday 12 July 1996.

New York police announced today the arrest of Frederick J. Thompson, head of
Jay Street Imports Inc., on charges of drug smuggling. Thompson was taken
from his Manhattan apartment in the early hours yesterday. His attorney,
Robert Giuliani, issued a statement denying any involvement with narcotics
on the part of his client. ``No way did Fred ever have dealings with dope'',
Guliani said.

A Jay Street
spokesperson said the company had ceased trading as of today. The company, a
medium-sized import-export concern established in 1989, had been the main
contractor in several collaborative transport ventures involving
Latin-American produce. Several associates of the firm moved yesterday to
distance themselves from the scandal, including the mid-western transportation
company Downing-Jones.

Thompson is understood to be accused of importing heroin into the United
States.
\end{quote}
\renewcommand{\baselinestretch}{1.4}
\small \normalsize
From this IE might produce information such as the following
(in some format to be determined according to user requirements,
e.g. SQL statements addressing some database schema).

First, a list of
entities and associated descriptive information. Relations of property type
are made explicit. Each entity has an id, e.g. {\tt ENTITY-2}, which can be 
used for cross-referencing between entities and for describing events 
involving entities. Each also has a type, or category, e.g. {\tt company},
{\tt person}. Additionally various type-specific information is available,
e.g., for dates, a {\tt normalisation} giving the date in standard format.

\renewcommand{\baselinestretch}{0.8}
\small \normalsize
\begin{verbatim}
    Reuter
            id:             ENTITY-1
            type:           company
            business:       news
    
    New York
            id:             ENTITY-2
            type:           location
            subtype:        city
            is_in:          US
    
    Wednesday 12 July 1996
            id:             ENTITY-3
            type:           date
            normalisation:  12/07/1996
    
    New York police
            id:             ENTITY-4
            type:           organisation
            location:       ENTITY-2
    
    Frederick J. Thompson
            id:             ENTITY-5
            type:           person
            aliases:        Thompson; Fred
            domicile:       ENTITY-7
            profession:     managing director
            employer:       ENTITY-6
    
    Jay Street Imports Inc.
            id:             ENTITY-6
            type:           organisation
            aliases:        Jay Street
            business:       import-export
    
    Manhattan
            id:             ENTITY-7
            type:           location
            subtype:        city
            is_in:          ENTITY-2
    
    Robert Guliani
            id:             ENTITY-8
            type:           person
            aliases:        Guliani

    1989
            id:             ENTITY-9
            type:           date
            normalisation:  ?/?/1989
    
    Latin-America
            id:             ENTITY-10
            type:           location
            subtype:        country
    
    Downing-Jones
            id:             ENTITY-11
            type:           organisation
            business:       transportation
    
    heroin
            id:             ENTITY-12
            type:           drug
            class:          A
    
    United States
            id:             ENTITY-13
            type:           location
            subtype:        country
\end{verbatim}
\renewcommand{\baselinestretch}{1.4}
\small \normalsize
(These results correspond to the combination of NE and TE
tasks; if we removed all but the type slots we would be left with the
NE data.)

Second, relations of event type, or scenarios:
\renewcommand{\baselinestretch}{0.8}
\small \normalsize
\begin{verbatim}
    narcotics-smuggling
            id:             EVENT-1
            destination:    ENTITY-13
            source:         unknown
            perpetrators:   ENTITY-5, ENTITY-6
            status:         on-trial
    
    joint-venture
            id:             EVENT-2
            type:           transport
            companies:      ENTITY-6, ENTITY-11
            status:         past
\end{verbatim}
(These results correspond to the ST task.)
\renewcommand{\baselinestretch}{1.4}
\small \normalsize

\section{\label{ml_ie}Multilingual IE}

The results described above
may then be translated for presentation to the user or for
storage in existing databases. In general this task is much easier than
translation of ordinary text, and is close to {\em
software localisation}, the process of making a program's
messages and labels on menus and buttons multilingual. Localisation involves
storing lists of direct translations for known items.
In our case these lists would store translations for
words such as ``entity'', ``location'', ``date'', ``heroin''. We also need
ways to display dates and numbers
in local formats, but code libraries are available for this type of problem.

Problems can arise where arbitrary pieces of text are used in the entity
description structures, for example the descriptor slot in MUC-6 TE objects.
Here a noun phrase from the text is extracted,
with whatever qualifiers, relative clauses etc.
happen to be there, so the language is completely unrestricted and
would need a full translation mechanism.

%
%

\newcommand{\etalchar}[1]{$^{#1}$}

\end{document}